\documentclass[10pt]{article}
\usepackage{times}
\usepackage{pgfplots}
\usepackage[a4paper, total={6in, 8in}]{geometry}
\usepackage{tikz}
\date{}

\usepackage{amsmath,amsfonts,bm}

\def\eqref#1{equation~\ref{#1}}

\def\1{\bm{1}}

\DeclareMathAlphabet{\mathsfit}{\encodingdefault}{\sfdefault}{m}{sl}
\SetMathAlphabet{\mathsfit}{bold}{\encodingdefault}{\sfdefault}{bx}{n}

\usepackage{url}

\usepackage{amsfonts}
\usepackage{microtype}
\usepackage{wrapfig}
\usepackage{times}
\usepackage{soul}
\usepackage{url}
\usepackage[hidelinks]{hyperref}
\usepackage[small]{caption}
\usepackage{graphicx}
\usepackage{amsmath}
\usepackage{amssymb}
\usepackage{booktabs}
\usepackage{algorithm}
\usepackage{algorithmic}
\usepackage{enumitem}

\usepackage{tikz}
\usepackage{pgfplots}
\pgfplotsset{compat=1.3}

\usepackage{natbib}
\usepackage{svg}

\usepackage{booktabs}
\usepackage{float}

\newcommand{\commentout}[1]{}
\title{Attribution-driven Causal Analysis for\\ Detection of Adversarial Examples }

\date{}

\author{Susmit Jha  \\
	Computer Science Laboratory \\
	SRI International  \\
		Menlo Park, CA \\
	\and 
	Sunny Raj, Steven Fernandes, Sumit Kumar Jha \\
	Computer Science Department\\ University of Central Florida \\
	Orlando, FL \\
	\and
	Somesh Jha\\
	Computer Science Department\\ University of Wisconsin \\
	Madison, WI \\
	\and
	Gunjan Verma, Brian Jalaian, Ananthram Swami\\
	U.S. Army Research Laboratory \\
		Adelphi, MD\\
	}

\begin{document}

\maketitle

\begin{abstract}
    Attribution methods  
have been developed to explain the decision of a machine learning model on a given input. 
We use the Integrated Gradient method for finding attributions 
to define the causal neighborhood  of an input by incrementally masking high attribution features. We study the robustness of machine learning models on  benign and adversarial inputs in this neighborhood. Our study indicates that  benign inputs are robust to the masking of high attribution features but  adversarial inputs generated by 
the state-of-the-art adversarial attack methods such as DeepFool, FGSM, CW and PGD, are not robust to such masking. Further, our study demonstrates  that this concentration of high-attribution features responsible for the incorrect decision is more pronounced in physically realizable adversarial examples. 
This difference 
in attribution of benign and adversarial inputs can be used to detect adversarial examples. Such a defense approach is independent of training data and attack method, and we demonstrate its effectiveness on digital and physically realizable perturbations. 

\end{abstract}

\section{Introduction}
\label{introduction}

Deep learning models have been wildly successful in applications such as computer vision~\citep{Krizhevsky:2017:ICD:3098997.3065386}, natural language processing, speech recognition, and automatic control. These models have reached human-level performance on several benchmarks~\citep{he2015delving,xiong2017toward}. But the adoption of these models in safety-critical or high-security applications is inhibited due to two major concerns: their brittleness to adversarial attack methods that can make imperceptible modification to inputs and trigger wrong decisions~\citep{szegedy2013intriguing,papernot2016cleverhans}, and the lack of interpretability~\citep{gunning2017explainable}. Significant progress has also been made towards adversarial robustness~\citep{papernot2016distillation,madry2017towards,engstrom2018evaluating} and explainability~\citep{li2015visual,yi2016lift,sundararajan2017axiomatic}, 
and a few recent theoretical studies~\citep{kilbertus2018generalization,chalasani2018adversarial} indicate a strong 
connection between these two issues.

In this paper, we investigate this connection between the resilience to adversarial perturbations and the attribution-based explanation of individual decisions by a machine learning model. Our study is motivated by 
Kahneman's decomposition~\citep{kahneman2011thinking} 
of cognition into two layers: System 1, or the intuitive system, and System 2, or the deliberate system. The original machine learning model represents the System 1 and the attribution analysis presented in this paper is the deliberative System 2 that can detect inconsistencies in System 1 by analyzing the attribution generated by the model on the input. Our central hypothesis  is that adversarial inputs are imperceptibly similar to benign inputs and are still able to trigger wrong decisions because of a  relatively small number of features with high attribution in the machine learning model. This concentration of  causal attributions is central to their physical realizability and perceptual indistinguishably from benign inputs.

\begin{wrapfigure}{r}{0.52\textwidth} 
\vspace{-0.42cm}
{\includegraphics[width=.14\textwidth]{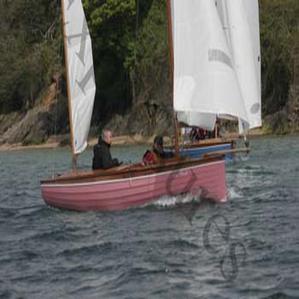}\hfill
\includegraphics[width=.14\textwidth]{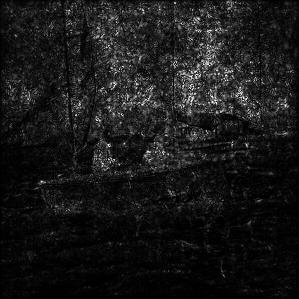}\hfill
\includegraphics[width=.14\textwidth]{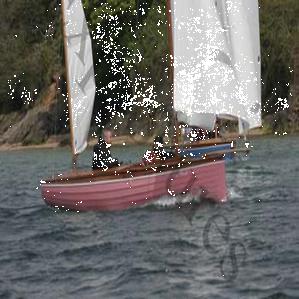}\\

\includegraphics[width=.14\textwidth]{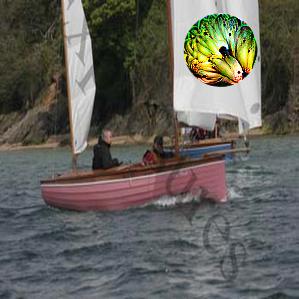}\hfill
\includegraphics[width=.14\textwidth]{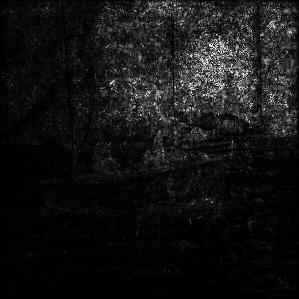}\hfill
\includegraphics[width=.14\textwidth]{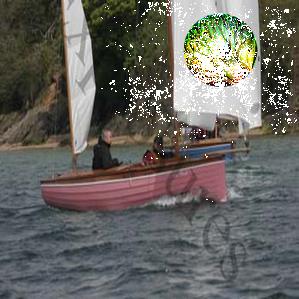}
}
\caption{The top row shows an original yawl image and the bottom row shows the adversarial input obtained by adding an adversarial patch~\citep{brown2017adversarial} which causes misclassification. The leftmost image is the input, next to it is the saliency map that shows the
 features (pixels) with positive attributions~\citep{szegedy2016rethinking}, and the rightmost  image shows the pixels with top 2\% attribution masked.
People in the boat are in the top 2\% positive attribution features in original image but not in the adversarial image.
The saliency map showing attributions are obtained using Integrated Gradient method~\citep{sundararajan2017axiomatic}
 \vspace{-0.45cm}
 }
\label{fig:motive}
\end{wrapfigure}
The example in Figure~\ref{fig:motive}
demonstrates the shift in positive attributions and the corresponding qualitative change in saliency map due to the adversarial perturbation.
We observe a similar shift in features with high magnitude attributions for the adversarial examples generated by the state-of-the-art purely digital attack methods
 such as DeepFool~\citep{moosavi2016deepfool}, 
projected gradient descent (PGD)~\citep{madry2017towards}, 
fast gradient sign method (FGSM)~\citep{goodfellow2014explaining} and
Carlini-Wagner (CW)~\citep{carlini2017towards}.

Adversarial perturbations 
often create a small fraction of high-attribution features responsible for the change in the output model without visible change in the input. Consequently, it is possible to identify an adversarial example by examining inputs in its causal neighborhood obtained by incrementally masking the features which have high magnitude attributions. While benign inputs are robust and the model does not change its decision on the input in this neighborhood, the decision of the model is not robust on the adversarial examples.
The localized nature of existing  physically realizable attacks naturally concentrates the high attribution features. The ease of detection of these adversarial examples suggests a trade off in physical realizability and indistinguishability in attribution space.
We make the following novel contributions in this paper:
\begin{itemize}[leftmargin=*]
  \item We define a causal neighborhood of an input in the attribution space that can be used to measure the efficacy of an adversarial attack  beyond perceptual similarity or the commonly used $L_p$ norms. Effective attack methods should not just be able to fool the intuitive System 1 i.e. the machine learning model but also the deliberative System 2 i.e. the attribution analysis presented here.
    \item We use  attribution methods to analyze the robustness of benign as well as adversarial inputs by incrementally masking high attribution features. Our empirical analysis includes digital  attack methods such as DeepFool, FGSM, CW and PGD, and physically realizable perturbations. 
    \item We propose a defense layer based on Kahneman's decomposition of cognition into intuitive System 1 and deliberative System 2.  This approach does not rely on analyzing training data such as  
    manifold-based defense~\citep{ilyas2017robust,jha2018detecting}, or statistical signature of the machine learning models such as logit pairing~\citep{engstrom2018evaluating},  or methods that exploit the knowledge of specific attack for adversarial training~\citep{tramer2017ensemble}, or
    robust optimization with $L_p$ norm bounds~\citep{madry2017towards,raghunathan2018certified}. 
\end{itemize}

The rest of the paper is organized as follows. We discuss relevant background and review related work in Section~\ref{sec:bg}.  We present our analysis approach and derived defense against adversarial perturbation in 
Section~\ref{sec:approach}. The experimental results from our empirical analysis on MNIST and ILSVRC datasets using
different attack methods are presented in Section~\ref{sec:exp}. We conclude in Section~\ref{sec:conc} by discussing the limitations of our approach, and the salient findings of our study. 

\vspace{-0.2cm}

\section{Background and Related Work}
\label{sec:bg}
~\citep{szegedy2013intriguing} first introduced adversarial examples using an L-BFGS method. These generated adversarial images could be generalized to different training databases and models; however they’re very rarely seen in test images. L-BFGS attack uses an expensive linear search method to obtain the optimal value, but this method is time-consuming and impractical. ~\citep{goodfellow2014explaining} proposed a fast gradient sign method (FGSM) to generate faster adversarial images as compared to the L-BFGS method; this method performed only a one-step gradient update at each pixel along the direction of the gradient’s sign. ~\citep{rozsa2016adversarial} replaced the sign of the gradients with raw gradients and proposed a new method called “fast gradient value method”. ~\citep{dong2018boosting} applied momentum to FGSM, with adversarial images generated more iteratively. The effectiveness and transferability improved by introducing momentum and by applying the ensembling method and the one-step attack. ~\citep{kurakin2016adversarial} further extended FGSM to a targeted attack, where the probability of the target class was maximized. This attack is also known as the one-step target class method.

Adversarial examples~\citep{goodfellow2014explaining} which produce incorrect decision from machine learning models 
can be obtained using a number of 
techniques~\citep{moosavi2016deepfool,madry2017towards,carlini2017towards}
which are now available in tools such
as Cleverhans~\citep{papernot2016cleverhans}.
While these methods rely on diffused digital transformation of the input, 
physically realizable attacks in form of patches or stickers that can be added to an image have also been developed. Adversarial patch ~\citep{brown2017adversarial} and localized and visible adversarial noise(LaVAN) ~\citep{lavan} methods  generate
patches that are universal and can be used to attack any image and are targeted to force classifier to output a particular output label. Their goal is  to generate universal noise “patches”
that can be physically printed and put on any image, in either a black-box (when the attacked network is unknown) or
white-box (when the attacked network is known) setup. 
A related attack is proposed in ~\citep{karmon2018lavan} to investigate the blind-spots of 
state-of-the-art image classifiers, and the kinds of noise that
can cause them to misclassify.
More diffused universal 
adversarial 
digital perturbations~\citep{moosavi2017universal}  
have also been proposed that can be applied to any input. Further, the adversarial 
examples have been shown to  transfer~\citep{liu2016delving,papernot2017practical} 
across models making them agnostic to
availability of model parameters and effective against even ensemble approaches. 
In addition to these inference-time attacks, machine learning models have also been shown to be susceptible to data poisoning attacks during training. 
Last few years of efforts to defend against  adversarial attacks have met with limited success.
While empirical approaches such as 
logit pairing~\citep{engstrom2018evaluating}  
and defensive distillation~\citep{papernot2016distillation} have shown effectiveness against particular attack methods, more principled techniques such as robust optimization~\citep{raghunathan2018certified,zhang2019theoretically} 
are limited to 
perturbations with bounded 
$L_p$ norm. This arms race between attacks and defenses are not limited to computer vision, but are also applicable to audio and natural language processing. 

A number of attribution techniques have been recently proposed in literature that assign positive and negative importance to an input feature for a given decision of the machine learning model. Attributions can correspond to the actual decision or to counterfactuals. Many prominent attribution methods are based on the gradient of the predictor function with respect to the 
input~\citep{simonyan2013deep,selvaraju2017grad,sundararajan2017axiomatic}. These approaches have been extended to counterfactual analysis~\citep{datta2016algorithmic} 
which has roots in cooperative game theory and revenue division. Counterfactual analysis is difficult when the number of possible output labels is large. We restrict our study to analysis of the actual attribution. 
Different attribution methods are compared in
~\citep{adebayo2018sanity}. 
The sensitivity of these attributions to perturbations in the input are studied in 
~\citep{ghorbani2017interpretation} and adversarial attacks are presented to create perceptively indistinguishable images with the same prediction label but  different attributions. Their results indicate that both System 1 and System 2 in our proposed approach can be independently attacked.

\begin{figure*}[t]
\centering
\includegraphics[width=14cm]{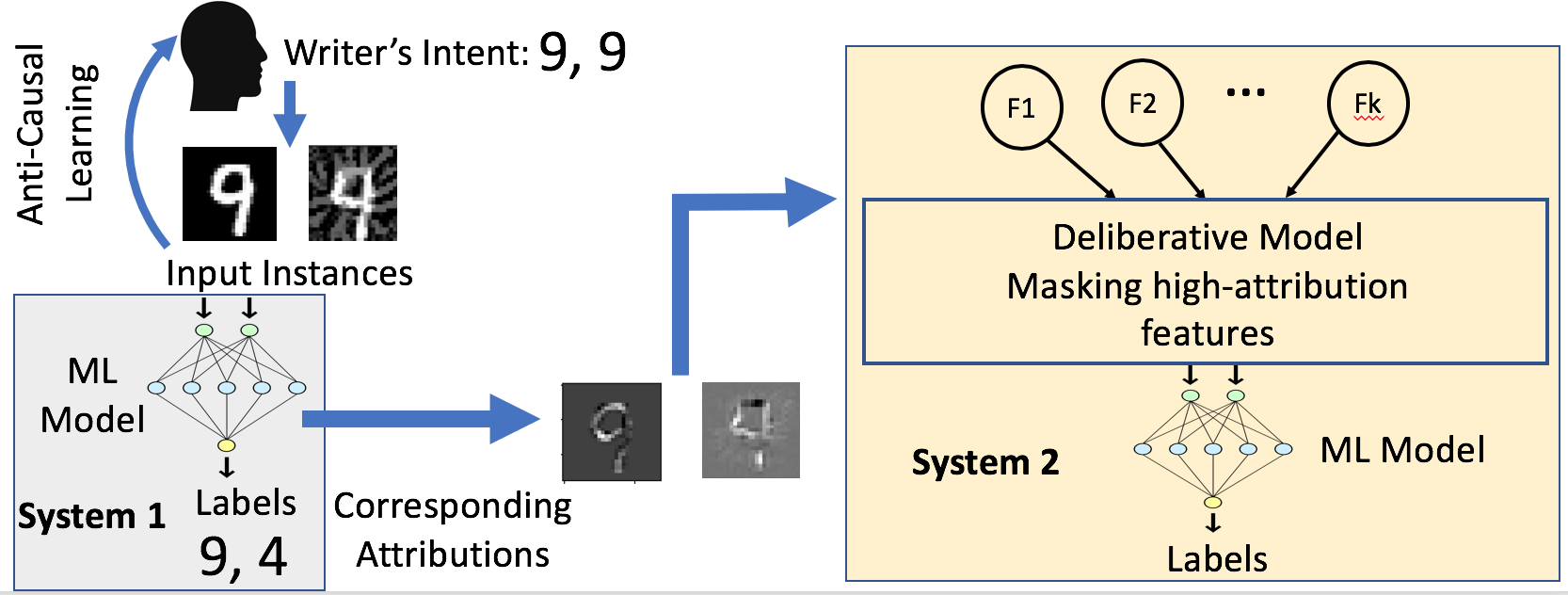}
\caption{The architecture of the proposed approach motivated by the two level Kahneman's decomposition of cognition. Typical machine learning models for  classification perform anti-causal learning  to determine the label from the input instance. As noted by ~\citep{chalasani2018adversarial}, such anti-causal reasoning lacks the natural continuity of causal mechanisms and is often not robust. But we view this model as System 1 and use attribution methods (Integrated Gradient in our experiments) to obtain features with positive and negative attributions. In this example with the MNIST dataset, we see that the adversarial perturbation that causes misclassification of 9 into 4 also significantly changes the attributions. For example, the top part of the perturbed 9 (misclassified as 4) has negative attribution. 
In deliberative System 2, we perform reasoning in the causal direction, and mask the high attribution features (pixels in this case) to obtain a number of input instances in the causal neighborhood of the original image. The original attributions are robust but the adversarial attributions are not robust which causes the model to assign different labels to images in the causal neighborhood of adversarial examples.   }
\vspace{-0.5cm}
\label{fig:overview} 
\end{figure*}

The goal of this paper is to combine both System 1 and System 2 to create a more resilient cognition model than any one of them alone. This is motivated by recent results in causal machine learning. 
~\citep{kilbertus2018generalization} argues that causal mechanisms are typically continuous but most learning problems such as classification are anti-causal. Strong generalization is limited in anti-causal learning and access to the causal model can remove such limitations.
Further, the lack of resilience to adversarial example is hypothesized to be the result of learning in anti-causal direction. We use the attributions and saliency map to identify the cause of the current decision of the machine learning model, and then construct a generator that can mask high-attribution features to explore the causal neighborhood of the input and observe model's behavior in this neighborhood. While learning is still in the anti-causal direction, we add a layer of causal deliberative System 2 that reasons in forward direction to evaluate the robustness of the obtained attribution. 

The connection between attributions and training with adversarial examples has also been 
recently studied~\citep{chalasani2018adversarial} where the authors note that adversarial training leads to feature concentration. They use the same attribution method of Integrated Gradients as our approach and report that adversarial training achieved much better feature concentration than traditional ${L}_1$ regularization. 
We study the connection between attribution and robustness from a different perspective. 
Our observation in this paper suggests that the features in adversarial inputs responsible for the wrong decision hinge on a relatively concentrated  small set of features and hence, are not robust to masking of features with high attribution magnitude.
Our result and the observation in 
~\citep{chalasani2018adversarial} are complementary. This suggests two parallel and independent approaches to increase robustness of a two-level cognition system. First, the System 1 comprising of the machine learning model can be made independently robust via adversarial training or direct regularization to 
concentrate the attributions, and second, System 2 can rely on masking high attribute features to check the robustness of the model's explanation for a specific instance. 

\section{Approach}
\label{sec:approach}
Figure~\ref{fig:overview}
shows the overall architecture of the proposed approach. 
The anti-causal learning used in tasks such as classification are often not robust to adversarial changes~\citep{chalasani2018adversarial}. Such models can be viewed as intuitive System 1 that makes fast  decisions. While these systems can definitely be made more robust and resilient to adversarial 
methods~\citep{madry2017towards,raghunathan2018certified}, our focus is on analyzing the attributions generated from the ML models on the inputs to detect their adversarial nature. We use image benchmarks as inputs and pixels as the features. But our approach can be adapted to other application domains and to more higher-level abstract features.  As discussed earlier, there are several recently proposed attribution 
methods that can be used to obtain the positive and negative attribution of different features for a given decision of the machine learning model. For a given neural network machine learning model $M$ and input $x$, the attribution for the model output $M(x)$ is computed for the features 
$F_1, F_2, \ldots, F_k$ of the input. 

As recognized in recent comparison of feature-attribution methods~\citep{adebayo2018sanity}, several feature attribution methods have quirks which obscure the impact of features on the model output, and evaluation of different attribution methods is challenging. 
In a recent paper~\citep{sundararajan2017axiomatic}, the authors identify several axioms that must be satisfied by a sound attribution method, and in particular propose a specific method that satisfies these axioms, which they call Integrated Gradients. We adopt this method in our study. The attribution in Integrated Gradient uses the notion of a baseline input. For example, the all dark image can be selected as a baseline for attribution over images. The attribution is then defined as the path integral of the gradients along the straight-line path from the baseline $x'$ to the input $x$, that is, the attribution for the $i$-th
feature is 
$$ IG(F_i(x)) = (x_i - x_i') \times \int_{\alpha=0}^1 \partial_i M(x'+\alpha(x-x')) d\alpha$$
where $\partial_iM(\cdot)$ denotes the gradient of $M(\cdot)$ along the $i$-th feature dimension. 

We adopt a simple masking model for causal generation of new inputs from the features. Given an input $x$ with baseline $x'$, the masking generator $G$ can select a subset of $n \leq k$ features $M = \{m_1, m_2, \ldots m_n\}$ to be masked and generate a new input $G(x,M)$ such that 
$F_i(G(x,M)) = F_i(x)$ for $i \not \in M$ and $F_i(G(x,M)) = F_i(x')$ for $i  \in M$. For example, if the baseline is a  dark image and the features are pixels of the image, the masking model selects a set of pixels $M$ of an input image and makes them dark. 
We use the attribution over features and this masking model  to define a causal neighborhood of the input $x$ as follows:
\vspace{0.1cm}

\noindent \textbf{Definition:} Given an input $x$ with features $F_1, F_2, \ldots, F_k$ and the corresponding attributions $IG(F_1(x)),$ $IG(F_2(x)),$ $\ldots, IG(F_k(x))$, we sort the features in decreasing order of their attribution $F_{j_1}, F_{j_2}, \ldots, F_{j_k}$ such that the attributions $IG(F_{j_1}(x)) \geq$  $IG(F_{j_2}(x))$ $\geq \ldots \geq  IG(F_{j_k}(x))$. We define a $\delta$ causal neighborhood of the input $x$ for $0\leq \delta \leq 1$ by constructing new inputs $G(x,M)$ obtained via masking features $M \subseteq \{F_{j_1}, F_{j_2}, \ldots, F_{j_{\lceil \delta k \rceil}} \}$, that is,
$${\cal{N}}_\delta(x) = \{ G(x,M) | M \subseteq \{F_{j_1}, F_{j_2}, \ldots, F_{j_{\lceil \delta k \rceil} } \} \} $$
We say that an input (benign or adversarial) is $\delta$ robust with respect to its attributions if the machine learning model produces the same output on all inputs in ${\cal{N}}_\delta(x)$. We restrict our study to positive attributions. Negative attributions can be used for counterfactual robustness testing but such a counterfactual testing will need to be done with respect to all alternate decisions. We denote this restricted neighborhood as ${\cal{N}}^+_\delta(x)$ where all features $F_{j_i}$ in $M$ are guaranteed to have positive attributions, that is,  $IG(F_{j_i}(x^M)) > 0$ for all $x^M \in {\cal{N}}^+_\delta(x)$. The following theorem follows from the monotonicity of $M$ with respect to the inclusion of features  positive attributes:
\vspace{0.1cm}

\noindent \textbf{Theorem:} If the outputs of a machine learning model $M$ on an input $x$ and for some input in its causal neighborhood $x^M \in N^+_\delta(x)$ are different, then the model $M$ also produces different outputs on the input $x$ and $G(x,M_\delta)$ where 
$M_\delta = \{F_{j_1}, F_{j_2}, \ldots, F_{j_{\lceil \delta k \rceil} } \}$. 

This monotonicity enables us to check the robustness of a machine learning model on an input in its causal neighborhood by just considering the farthest input. If the model produces different output, then it is not robust on this input. Algorithm 1 summarizes this robustness check procedure which forms the core of the deliberative System 2.

\begin{algorithm}[tb]
\caption{Evaluate robustness $\rho$ of machine learning model $M$ on input $x$ with granularity $\epsilon$ 
}
\label{alg:attack1}
\begin{algorithmic}[1]
\STATE Using Integrated Gradient, compute the attributions of the features $F_{1}, F_{2}, \ldots, F_{k}$ of $x$ as  $IG(F_1(x)), IG(F_2(x)), \ldots, IG(F_k(x))$ 
\STATE Sort the features in decreasing order of attributions $F_{j_1}, F_{j_2}, \ldots, F_{j_k}$ such that the attributions $IG(F_{j_1}(x)) \geq  IG(F_{j_2}(x)) \geq \ldots \geq  IG(F_{j_k}(x))$.
\STATE Filter the features to keep only those with positive attributions $F_{j_1}, F_{j_2}, \ldots, F_{j_m}$ such that $IG(F_{j_1}(x)) \geq  IG(F_{j_2}(x)) \geq \ldots \geq  IG(F_{j_m}(x)) > 0$.
\STATE $\delta \leftarrow \epsilon$
\STATE $M_\delta \leftarrow \{F_{j_1}, F_{j_2}, \ldots, F_{j_{\lceil \epsilon m \rceil} } \}$
\WHILE {$M$ produces same outputs on $x$ as well as on $G(x,M_\delta)$, and $\delta \leq 1$}
\STATE $\delta \leftarrow \delta + \epsilon$
\STATE $M_\delta \leftarrow \{F_{j_1}, F_{j_2}, \ldots, F_{j_{\lceil \delta m \rceil} } \}$
\ENDWHILE
\RETURN $\rho(M,x) = \delta$
\end{algorithmic}
\end{algorithm}

We can lift the computation of the robustness of machine learning model $M$ on input $x$ to define the attribution sensitivity on a dataset $X$ as follows: 
$$ {\cal{S}}(M,X,\delta) = 1 - \frac{|\{x | x \in X \; \text{and} \; \rho(M,x) \leq \delta \}|}{|X|} $$
where $|\cdot|$ denotes the size of the set.
The central goal of the paper is to study the connection between attributions and adversarial robustness, and to argue the need for Kahneman's System 1 and System 2 cognition for more robust perception. While such an approach will not create a fool-proof defense against adversarial examples, it is an essential first-step towards building trusted high assurance intelligent agents. We accomplish this
by computing the attribution sensitivity on original datasets and adversarially perturbed datasets. An effective adversarial attack must have smaller attribution sensitivity similar to original inputs to avoid detection.

.
\section{Analysis Results}
\label{sec:exp}
We perform the attribution sensitivity analysis on three datasets to test our hypothesis that adversarial examples are less robust (more sensitive) to masking in the attribution space when compared to the robustness of original images to such masking:
\begin{itemize}
    \item MNIST images and adversarial datasets obtained using FGSM (with different epsilon bounds), DeepFool, PGD and CW attack methods.
    \item ImageNet images and adversarial datasets obtained using FGSM (with different epsilon bounds), DeepFool, PGD and CW attack methods.
    \item Physically realizable adversarial patch~\citep{brown2017adversarial} and LaVAN \citep{lavan} attack applied to 1000 images from ImageNet. For adversarial patch attack we used a patch size of 25\%for two patch types:  banana and toaster. For LaVAN we used baseball patch of size $50\times50$ pixels. 
\end{itemize}

\begin{figure*}[htbp]
\centering 
\begin{tikzpicture}
    \footnotesize
    \begin{axis}[width=0.45\textwidth, xlabel=Percentage of Top Attributions Masked (100 $\times$ $\delta$), ylabel near ticks,xmajorgrids,ymajorgrids,ymin=0,ymax=80, ylabel style={align=center}, ylabel=Percentage of Changed Labels\\  (100 $\times$  ${\cal{S}}$) \\, legend columns=-1, legend to name=named]
    \addplot[smooth,mark=*,blue] plot coordinates {
        (0,0.0)
        (1,0.0)
        (2,0.01)
        (3,0.14)
        (4,0.43)
        (5,0.94)
        (6,1.66)
        (7,2.51)
        (8,3.55)
        (9,4.7)
        (10,5.92)
        (11,7.18)
        (12,8.41)
        (13,9.74)
        (14,10.98)
        (15,12.33)
        (16,13.52)
        (17,14.8)
        (18,16.03)
        (19,17.36)
        (20,18.63)
        (21,19.69)
        (22,20.86)
        (23,22.07)
        (24,23.21)
        (25,24.52)
    };
    \addlegendentry{Original Images}
     \addplot[smooth,mark=square,cyan] plot coordinates {
        (0,0.0)
        (1, 21.22)
        (2, 42.36)
        (3, 52.22)
        (4, 58.14)
        (5, 61.83)
        (6, 64.30)
        (7, 65.87)
        (8, 67.26)
        (9, 68.20)
        (10, 68.71)
        (11, 68.71)         
        (13, 69.27)
        (14, 69.59)
        (15, 71.06)
        (16, 71.13)
        (17, 71.13)        
        (20, 72.62)
        (21, 72.63)
        (22, 72.74)
        (23, 72.88)
        (24, 72.97)
        (25, 73.15)
    };
    \addlegendentry{FGSM2,5,10}

      \addplot[color=green,mark=o] plot coordinates {
        (0,0.0)
        (1,2.11)
        (2,27.13)
        (3,38.18)
        (4,45.43)
        (5,50.09)
        (6,53.57)
        (7,56.13)
        (8,58)
        (9,59.47)
        (10,60.76)
        (11,61.9)
        (12,62.78)
        (13,63.58)
        (14,64.26)
        (15,64.8)
        (16,65.32)
        (17,65.8)
        (18,66.27)
        (19,66.71)
        (20,67.18)
        (21,67.45)
        (22,67.75)
        (23,68.14)
        (24,68.4)
        (25,68.91)
        };
    \addlegendentry{DeepFool}
    \addplot[color=purple,mark=triangle]
        plot coordinates {
                (0,0.0)
        (1,26.96)
        (2,33.00)
        (3,36.09)
        (4,37.05)
        (5,39.07)
        (6,40.26)
        (7,40.63)
        (8,41.77)
        (9,42.89)
        (10,43.29)
        (11,43.29)         
        (14,44.01)
        (15,48.17)
        (16,48.61)
        (17,48.61)         
        (20,52.56)
        (21,52.69)
        (22,53.62)
        (23,54.18)
        (24,54.65)
        (25,55.25)
        };
    \addlegendentry{Carlini Wagner}
   \addplot[color=orange,mark=+]
        plot coordinates {
        (0,0.0)
        (1,7.90)
        (2,20.76)
        (3,30.45)
        (4,38.40)
        (5,45.04)
        (6,49.84)
        (7,53.73)
        (8,56.59)
        (9,58.81)
        (10,60.71)
        (11,62.12)
        (12,63.29)
        (13,64.39)
        (14,65.28)
        (15,66.14)
        (16,66.72)
        (17,67.72)       
        (19,68.11)
        (20,68.53)
        (21,68.84)
        (22,69.09)
        (23,69.34)
        (24,69.54)
        (25,69.81)
        };
    \addlegendentry{PGD}
     
    \end{axis}
\end{tikzpicture}
\quad
\begin{tikzpicture}
    \footnotesize
    \begin{axis}[
        width=0.45\textwidth,
        xlabel=Percentage of Top Attributions Masked (100 $\times$ $\delta$), ylabel near ticks, xmajorgrids,ymajorgrids,ymin=0,ymax=100,
        ylabel style={align=center}, ylabel=Percentage of Changed Labels\\ \;(100 $\times$  ${\cal{S}}$), legend columns=-1,
                legend to name=named1]
         \addplot[smooth,mark=*,blue] plot coordinates {
        (0.0,0.0)
        (0.02,7.8)
        (0.04,11.1)
        (0.06,13.6)
        (0.08,16.4)
        (0.1,18.2)
        (0.12,19.1)
        (0.14,20.5)
        (0.16,22.0)
        (0.18,23.5)
        (0.2,23.9)
        (0.22,24.6)
        (0.24,25.4)
        (0.26,26.5)
        (0.28,27.5)
        (0.3,28.4)
        (0.32,29.4)
        (0.34,30.1)
        (0.36,29.5)
        (0.38,31.1)
        (0.4,32.2)
    };
    \addlegendentry{Original Images}
    \addplot[smooth,mark=square,cyan] plot coordinates {
        (0.0,0.0)
        (0.02,31.29)
        (0.04,39.71)
        (0.06,44.71)
        (0.08,49.43)
        (0.1,53.57)
        (0.12,56.14)
        (0.14,58.14)
        (0.16,59.71)
        (0.18,62.0)
        (0.2,64.57)
        (0.22,67.29)
        (0.24,68.71)
        (0.26,69.57)
        (0.28,70.0)
        (0.3,71.71)
        (0.32,72.71)
        (0.34,74.0)
        (0.36,74.71)
        (0.38,74.71)
        (0.4,76.14)
    };
    \addlegendentry{FGSM2}

    \addplot[smooth,color=red,mark=x]
        plot coordinates {
        (0.0,0.0)
        (0.02,27.86)
        (0.04,38.22)
        (0.06,43.34)
        (0.08,45.63)
        (0.1,49.26)
        (0.12,51.82)
        (0.14,55.19)
        (0.16,57.34)
        (0.18,59.49)
        (0.2,61.37)
        (0.22,62.72)
        (0.24,64.20)
        (0.26,65.41)
        (0.28,66.62)
        (0.3,67.43)
        (0.32,68.64)
        (0.34,68.91)
        (0.36,70.39)
        (0.38,70.39)
        (0.4,71.20)
        };
    \addlegendentry{FGSM5}
    
    \addplot[smooth,color=brown,mark=diamond]
        plot coordinates {
        (0.0,0.0)
        (0.02,32.21)
        (0.04,42.42)
        (0.06,47.92)
        (0.08,51.28)
        (0.1,54.36)
        (0.12,55.70)
        (0.14,57.32)
        (0.16,60.13)
        (0.18,62.28)
        (0.2,63.49)
        (0.22,65.50)
        (0.24,66.71)
        (0.26,68.19)
        (0.28,68.46)
        (0.3,70.20)
        (0.32,71.95)
        (0.34,72.35)
        (0.36,72.89)
        (0.38,74.23)
        (0.4,75.30)
        };
    \addlegendentry{FGSM10}

    \addplot[color=green,mark=o]
        plot coordinates {
        (0.0,0.0)
        (0.02,94.8)
        (0.04,95.9)
        (0.06,96.2)
        (0.08,95.5)
        (0.1,95.9)
        (0.12,96.1)
        (0.14,95.7)
        (0.16,96.3)
        (0.18,96.0)
        (0.2,96.2)
        (0.22,96.6)
        (0.24,96.5)
        (0.26,96.7)
        (0.28,96.6)
        (0.3,96.7)
        (0.32,97.1)
        (0.34,97.1)
        (0.36,96.9)
        (0.38,96.9)
        (0.4,97.1)
        };
    \addlegendentry{Deep Fool}
    
    \addplot[color=purple,mark=triangle]
        plot coordinates {
        (0.0,0.0)
        (0.02,80.4)
        (0.04,86.6)
        (0.06,89.6)
        (0.08,90.9)
        (0.1,92.1)
        (0.12,92.0)
        (0.14,92.5)
        (0.16,93.8)
        (0.18,94.0)
        (0.2,94.1)
        (0.22,93.8)
        (0.24,94.8)
        (0.26,95.6)
        (0.28,95.3)
        (0.3,95.6)
        (0.32,95.4)
        (0.34,95.6)
        (0.36,95.2)
        (0.38,95.3)
        (0.4,95.9)
        };
    \addlegendentry{CW}
   
   \addplot[color=orange,mark=+]
        plot coordinates {
        (0.0,0.0)
        (0.02,35.87)
        (0.04,48.70)
        (0.06,51.30)
        (0.08,57.31)
        (0.1,61.32)
        (0.12,62.12)
        (0.14,62.93)
        (0.16,63.33)
        (0.18,64.13)
        (0.2,65.93)
        (0.22,66.33)
        (0.24,66.13)
        (0.26,66.73)
        (0.28,67.33)
        (0.3,66.93)
        (0.32,68.14)
        (0.34,68.74)
        (0.36,70.94)
        (0.38,71.14)
        (0.4,71.34)
        };
    \addlegendentry{PGD}
   
    \end{axis}
   
\end{tikzpicture}    
\ref{named1}
\caption{ (Left) Percentage of changed labels and percentage of masked attributions for adversarial attacks FGSM, DeepFool, CW, and PGD on MNIST dataset. (Right) Percentage of changed labels and fraction of masked pixels on ImageNet database. FGSM attacks with different $\epsilon=2, 5, 10$ are evaluated for ImageNet. }
\label{fig:mnistimagenet}
\end{figure*}
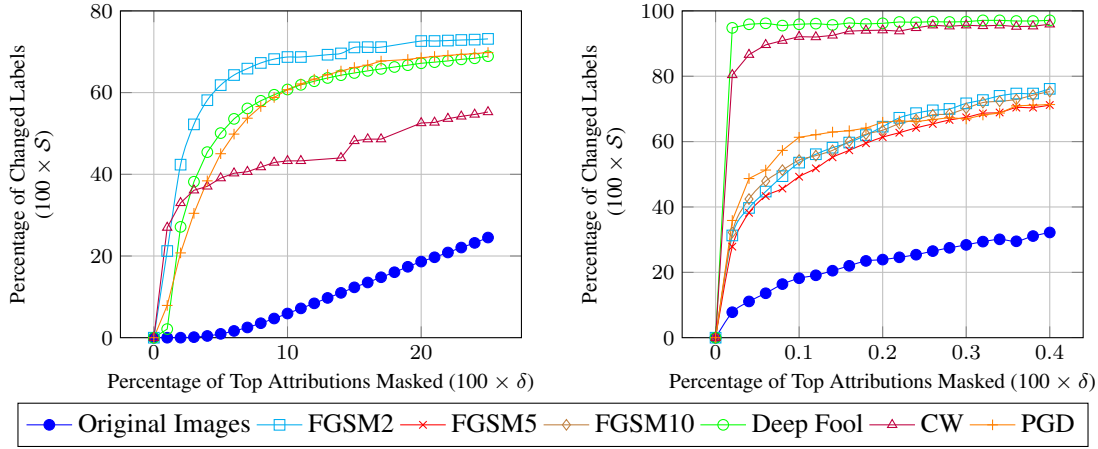

\subsection{MNIST Dataset and Digital Attacks}
The original MNIST images show attribution robustness where removing the top $10\%$ attributions does not change the decision of the model on $94\%$ of the images. Even removing $20\%$ of the attributions does not cause change in the labels assigned to $81\%$ of the MNIST images. 
In contrast, masking the top $20\%$ attribution has a much  stronger impact on the adversarial images. As shown
in Figure \ref{fig:mnistimagenet}, this changes the assigned labels of adversarial images generated using   FGSM, DeepFool, CW and PGD by $72\%$, $67\%$, $53\%$, and $68\%$ respectively. 
\newcommand{\noop}[1]{}

\subsection{ImageNet Dataset and Digital Attacks}

The masking of a small number of high-attribution pixels does not cause a change in the labels of the original ImageNet images as shown in Figure~\ref{fig:mnistimagenet}. Labels assigned to $82\%$ of ImageNet images remain unchanged when pixels corresponding to top $0.1\%$ of attributions are masked. The fraction of unchanged labels remains at $67\%$ even when pixels corresponding to top $0.4\%$ of attributions are masked.  In contrast, $76\%,71\%$ and $75\%$ of the images changed label on masking $0.4\%$ of pixels corresponding to top attributions for the adversarial examples generated using FGSM with different epsilon values. This percentage for PGD$(\epsilon=2)$ was $71\%$, $97\%$ for CW and $95\%$ for DeepFool.

\noop{
\begin{figure}[H]
\centering
\begin{tikzpicture}
    \footnotesize
    \begin{axis}[
        xlabel=Percentage of Top Attributions Masked (100 $\times$ $\delta$), ylabel near ticks, xmajorgrids,ymajorgrids,ymin=0,ymax=100,
        ylabel=Percentage of Changed Labels\\ (100 $\times$  ${\cal{S}}$), legend columns=-1,
                legend to name=named]
    \end{axis}
\end{tikzpicture}
\caption{Percentage of correct labels and masked attributions for 1000 images from the ImageNet database.}
\label{fig:imagenet_original} 
\end{figure}
}

\subsection{ImageNet Dataset and Physical Attacks}

Using 1000 images from the ImageNet dataset, we created adversarial patch and LaVAN attacks for the InceptionV3 deep learning model. For adversarial patch attack we used a patch size of $25\%$ for two patch types: banana and toaster. A total of $347$ out of $1000$ images were successfully perturbed and assigned banana label on applying the banana patch. Corresponding number for the toaster patch was $376$. We observe that  there was a change in label for $99.71\%$ of images having the banana patch on masking top $0.4\%$ of high attribution pixels. Similarly, $98.14\%$ of images having the toaster patch changed label on masking top $0.4\%$ of high attribution pixels. For LaVAN attack we used baseball patch of size $50\times50$ pixels. A total of $747$ out of $1000$ images were successfully perturbed and assigned baseball label. We observe that $99.20\%$ of images having baseball patch changed label when we masked top $0.4\%$ of attributions. Detection percentages for various pixel masking percentages for images with banana, toaster and baseball patch are shown in Figure~\ref{fig:toasterBanana}.

\begin{figure}[htbp]
\begin{tikzpicture}
\footnotesize
    \begin{axis}[
       width=0.5\textwidth,
        xlabel=Percentage of Top Attributions Masked,xmajorgrids,ymajorgrids,ylabel near ticks,
        ylabel style={align=center}, ylabel=Percentage of Changed Labels, 
        legend columns=-1,
        legend style={at={(0,-0.25)},anchor=north west, cells={align=left}},
        legend columns=2
                ]
    
    \addplot[smooth,mark=*,blue] plot coordinates {
        (0.0,0.0)
        (0.02,7.8)
        (0.04,11.1)
        (0.06,13.6)
        (0.08,16.4)
        (0.1,18.2)
        (0.12,19.1)
        (0.14,20.5)
        (0.16,22.0)
        (0.18,23.5)
        (0.2,23.9)
        (0.22,24.6)
        (0.24,25.4)
        (0.26,26.5)
        (0.28,27.5)
        (0.3,28.4)
        (0.32,29.4)
        (0.34,30.1)
        (0.36,29.5)
        (0.38,31.1)
        (0.4,32.2)
    };
    \addlegendentry{Original}

    \addplot[smooth,mark=square*,red] plot coordinates {
        (0.0,0.0)
        (0.02,68.01)
        (0.04,85.30)
        (0.06,90.20)
        (0.08,93.66)
        (0.1,95.39)
        (0.12,96.83)
        (0.14,97.98)
        (0.16,98.56)
        (0.18,98.56)
        (0.2,99.14)
        (0.22,99.14)
        (0.24,99.14)
        (0.26,99.42)
        (0.28,99.71)
        (0.3,99.71)
        (0.32,99.71)
        (0.34,99.71)
        (0.36,99.71)
        (0.38,99.71)
        (0.4,99.71)
    };
    \addlegendentry{Banana Patch}

    \addplot[smooth,color=green,mark=triangle*]
        plot coordinates {
        (0.0,0.0)
        (0.02,46.28)
        (0.04,63.83)
        (0.06,73.94)
        (0.08,81.65)
        (0.1,86.44)
        (0.12,88.83)
        (0.14,90.96)
        (0.16,93.35)
        (0.18,94.15)
        (0.2,94.68)
        (0.22,95.74)
        (0.24,96.54)
        (0.26,97.07)
        (0.28,97.07)
        (0.3,97.60)
        (0.32,97.87)
        (0.34,97.87)
        (0.36,97.61)
        (0.38,98.14)
        (0.4,98.14)
        };
    \addlegendentry{Toaster Patch}

 \addplot[smooth,color=purple,mark=diamond*]
        plot coordinates {
        (0.0,0.0)
        (0.02,12.18)
        (0.04,18.47)
        (0.06,25.03)
        (0.08,36.01)
        (0.1,44.44)
        (0.12,54.22)
        (0.14,62.92)
        (0.16,72.02)
        (0.18,77.78)
        (0.2,82.73)
        (0.22,87.68)
        (0.24,91.03)
        (0.26,92.64)
        (0.28,95.72)
        (0.3,97.19)
        (0.32,97.72)
        (0.34,98.26)
        (0.36,98.53)
        (0.38,99.06)
        (0.4,99.20)
        };
    \addlegendentry{Baseball Patch}
    \end{axis}
    \hfill 
\end{tikzpicture}
\hfill 
\begin{tikzpicture}
    \footnotesize
    \begin{axis}[
        width=0.47\textwidth,
        xlabel=Percentage of Top Attributions Masked, ylabel near ticks, xmajorgrids,ymajorgrids,
        ylabel style={align=center}, ylabel=Percentage of Changed Labels,
                legend style={at={(-0,-0.25)},anchor=north west},
        legend columns=2
                        ]

    \addplot[smooth,mark=*,blue] plot coordinates {
        (0.0,0.0)
        (0.02,7.8)
        (0.04,11.1)
        (0.06,13.6)
        (0.08,16.4)
        (0.1,18.2)
        (0.12,19.1)
        (0.14,20.5)
        (0.16,22.0)
        (0.18,23.5)
        (0.2,23.9)
        (0.22,24.6)
        (0.24,25.4)
        (0.26,26.5)
        (0.28,27.5)
        (0.3,28.4)
        (0.32,29.4)
        (0.34,30.1)
        (0.36,29.5)
        (0.38,31.1)
        (0.4,32.2)
    };
    \addlegendentry{Original}
    
    \addplot[smooth,color=red,mark=square*] plot coordinates {
        (0.0,0.0)
        (0.02,68.01)
        (0.04,85.30)
        (0.06,90.20)
        (0.08,93.66)
        (0.1,95.39)
        (0.12,96.83)
        (0.14,97.98)
        (0.16,98.56)
        (0.18,98.56)
        (0.2,99.14)
        (0.22,99.14)
        (0.24,99.14)
        (0.26,99.42)
        (0.28,99.71)
        (0.3,99.71)
        (0.32,99.71)
        (0.34,99.71)
        (0.36,99.71)
        (0.38,99.71)
        (0.4,99.71)
    };
    \addlegendentry{Patch Size 25$\%$}

    \addplot[smooth,color=olive,mark=square*]
        plot coordinates {
        (0.0,0.0)
        (0.02,9.08)
        (0.04,15.43)
        (0.06,22.10)
        (0.08,27.79)
        (0.1,35.23)
        (0.12,39.82)
        (0.14,43.98)
        (0.16,47.48)
        (0.18,50.77)
        (0.2,54.81)
        (0.22,58.53)
        (0.24,61.16)
        (0.26,64.22)
        (0.28,67.29)
        (0.3,69.37)
        (0.32,72.10)
        (0.34,73.63)
        (0.36,75.82)
        (0.38,76.91)
        (0.4,79.65)
        };
    \addlegendentry{Patch Size 30$\%$}
    
    \addplot[smooth,color=red,mark=diamond*] plot coordinates {
        (0.0,0.0)
        (0.02,6.59)
        (0.04,12.64)
        (0.06,19.23)
        (0.08,24.29)
        (0.1,28.57)
        (0.12,33.08)
        (0.14,37.25)
        (0.16,41.10)
        (0.18,45.82)
        (0.2,50.33)
        (0.22,53.08)
        (0.24,57.47)
        (0.26,60.99)
        (0.28,63.74)
        (0.3,67.25)
        (0.32,69.56)
        (0.34,71.87)
        (0.36,74.29)
        (0.38,76.59)
        (0.4,79.01)
    };
    \addlegendentry{Patch Size 35$\%$}

    \addplot[color=purple,mark=triangle*] plot coordinates {
        (0.0,0.0)
        (0.02,7.31)
        (0.04,13.35)
        (0.06,19.39)
        (0.08,24.58)
        (0.1,30.61)
        (0.12,36.76)
        (0.14,44.07)
        (0.16,49.36)
        (0.18,54.24)
        (0.2,59.22)
        (0.22,63.67)
        (0.24,67.37)
        (0.26,71.08)
        (0.28,73.62)
        (0.3,76.48)
        (0.32,79.13)
        (0.34,81.36)
        (0.36,84.00)
        (0.38,84.75)
        (0.4,86.97)
    };
    \addlegendentry{Patch Size 40$\%$}

    \end{axis}
\end{tikzpicture}

\caption{(Left) Percentage of correct labels and masked attributions for 1000 images from the ImageNet database is shown in blue. Banana patch and toaster patch were generated using adversarial patch method~\citep{brown2017adversarial}. Baseball patch was generated using LaVAN method~\citep{lavan}. Dropping 0.4\% of the attribution causes 99.71\% of the attacks based on banana patches to be detected, 98.14\% of the attacks based on toaster patch to be detected and 99.20\% of the attacks based on baseball patch to be detected. (Right) Masking $0.4\%$ of attributions causes nearly $ 80\%$ of labels to change for images with adversarial patches. }
\label{fig:toasterBanana}
\end{figure}

\begin{figure}[htbp]
\centering
\includegraphics[width=.15\textwidth]{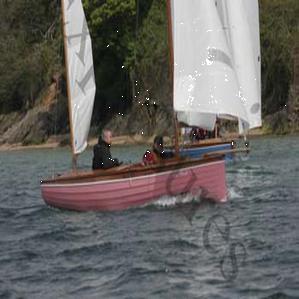}\hfill
\includegraphics[width=.15\textwidth]{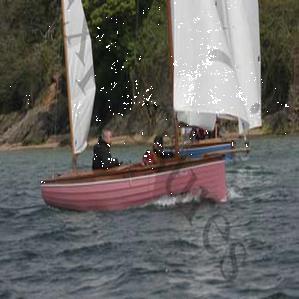}\hfill
\includegraphics[width=.15\textwidth]{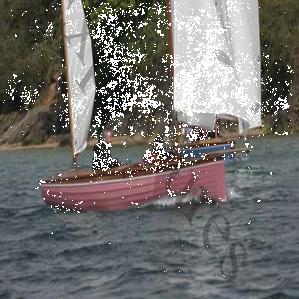} \hfill
\includegraphics[width=.15\textwidth]{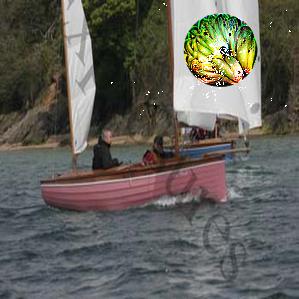}\hfill
\includegraphics[width=.15\textwidth]{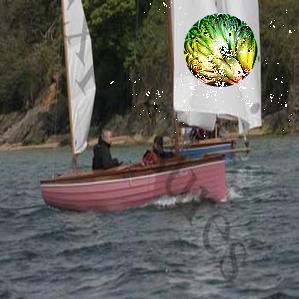}\hfill
\includegraphics[width=.15\textwidth]{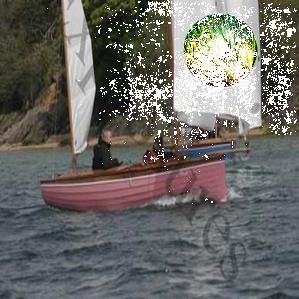} 

\caption{(From left to right) the original image with a label of yawl;  masking its top $0.2\%$ attribution;  masking its top $4\%$ of attribution; image with a banana patch generated using adversarial patch method; masking its top $0.2\%$ attribution; (rightmost) masking its top $4\%$ of attribution. Most pixels are masked from the noise patch leading to a change in label. Original images are robust to small pixel masking and retain their labels.}
\label{fig:physicalYawl}
\end{figure}

\begin{figure}[htbp]
\centering
\includegraphics[width=.15\textwidth]{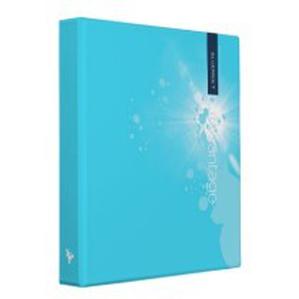}\hfill
\includegraphics[width=.15\textwidth]{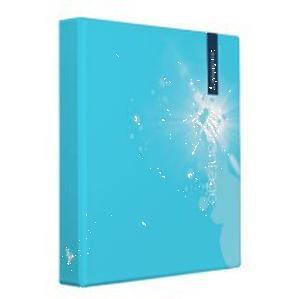}\hfill
\includegraphics[width=.15\textwidth]{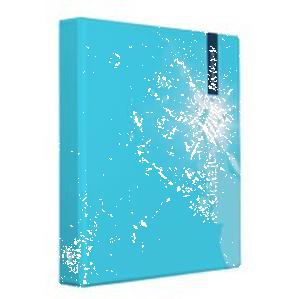}  \hfill
\includegraphics[width=.15\textwidth]{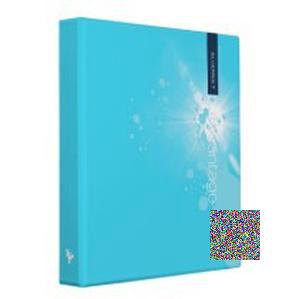}\hfill
\includegraphics[width=.15\textwidth]{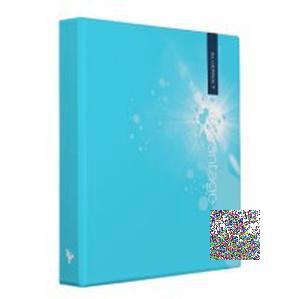}\hfill
\includegraphics[width=.15\textwidth]{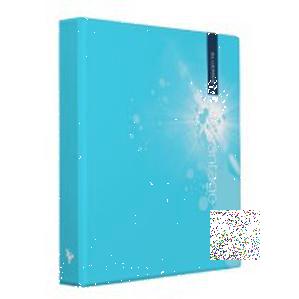}

\caption{(From left to right) original image; masking its top $0.2\%$ attribution; masking its top $2\%$ of attribution; image with a baseball patch generated using LaVAN method that causes the Inception deep neural net to label the entire image as a baseball; masking  its top $0.2\%$ attribution; (rightmost) masking top $2\%$ of its top attributions. Most pixels are again masked from the noise patch leading to a change in label. Original images are robust to small pixel masking and retain their labels.}
\label{fig:physicalBinder}
\end{figure}

We observe that most of the high attribution pixels in adversarial images is localized in the patch and masking these pixels leads to a change in the image label. An example of this behaviour can been seen for adversarial patch and LaVAN methods in Figures \ref{fig:physicalYawl} and \ref{fig:physicalBinder} respectively. In Figure,  \ref{fig:physicalYawl} most of the masking is localized around the banana patch; similarly,  most of the masking is localized around the baseball patch in Figure \ref{fig:physicalBinder}. 

A smaller size of adversarial patch is more desirable as an attacker can apply it to more surfaces. In our testing of adversarial patch method, $25\%$ was the smallest patch size that could fool Inception classifier consistently. We applied our method on various patch sizes ranging from $25\%$ to $40\%$ in increments of $5\%$ to test the robustness of our method. Results showing detection percentages for various patch sizes is presented in the right of Figure \ref{fig:toasterBanana}.
Smaller patches are relatively easier to detect due to high attribute feature concentration suggesting a trade off in making attacks physically realizable and yet imperceptible, and makings it robust in the attribution space.

\section{Conclusion}
\label{sec:conc}
We investigated the connection between the resilience to adversarial perturbations and the attribution-based explanation  of individual  decisions  generated by  machine  learning models. 
We observe that adversarial inputs are not robust in attribution space, that is, masking a few features with high attribution leads to change in decision of the machine learning model on the adversarial examples. In contrast, the natural inputs are robust in attribution space. 
This observation can be used to devise a new defense approach for machine learning models that uses attribution and incremental masking for filtering out adversarial examples. We propose such a deliberative masking based analysis approach as System 2 of Kahneman's two layer cognition system with the actual machine learning model serving as System 1. We demonstrate the effectiveness of our approach on a set of benchmarks that include diffused digital perturbations as well as physically realizable perturbations. The physically realizable adversarial perturbations are often localized; hence, they are easier to detect using attributions. 
Further, attribution-based defense is agnostic to the attack method and also independent of the training data. Instead, it relies on analysis of the result and the attribution generated by a machine learning model on a specific input. 

Our approach can make machine learning models more resilient to adversarial attacks because fooling this two layer cognition system requires not only attacking the original model but also ensuring that the attribution generated for the adversarial example is similar to the original examples. Both System 1 and System 2 must be simultaneously compromised  for a successful adversarial attack. Our experimental results indicate that the current adversarial attack methods are not able to achieve this. 
The adversarial inputs generated by the state-of-the-art attack methods such as DeepFool, FGSM, CW and PGD are able to fool a machine learning model but they are not robust in the attribution space. New attack methods must be evaluated by not only measuring the observed loss of accuracy but also evaluating their robustness in attribution space.

\bibliographystyle{iclr2019_conference}
\bibliography{paper.bib,kacpaper.bib}

\end{document}